\documentclass[10pt,twocolumn,letterpaper]{article}

\usepackage{cvpr}              % To produce the CAMERA-READY version

\usepackage{kotex}
\usepackage{amsmath}

\usepackage[dvipsnames]{xcolor}
\usepackage{algorithm}
\usepackage{algorithmic}
\usepackage{multirow}
\usepackage{booktabs}
\usepackage{amsmath}

\DeclareMathOperator*{\argmax}{arg\,max} 
\definecolor{cvprblue}{rgb}{0.21,0.49,0.74}
\usepackage[pagebackref,breaklinks,colorlinks,citecolor=cvprblue]{hyperref}

\def\confName{CVPR}
\def\confYear{2024}

\title{Practical Dataset Distillation Based on Deep Support Vectors %in Practical Settings
}
\author{
    Hyunho Lee \hspace{5em} Junhoo Lee \hspace{5em} Nojun Kwak\thanks{Corresponding author.} \\ \vspace{0.1mm}
    Seoul National University \\
    {\ttfamily\small \{hhlee822, mrjunoo, nojunk\}@snu.ac.kr}
}

\begin{document}
\maketitle
\begin{abstract}
%Conventional dataset distillation requires significant computational resources and assumes access to the entire dataset, an assumption impractical as it presumes all data resides on a central server. In this paper, we shift our focus to dataset distillation in practical scenarios where access is limited to only a fraction of the entire dataset. We propose a novel distillation method that enhances the conventional process by incorporating general model knowledge through the inclusion of DKKT loss. Our approach demonstrates improved performance in practical settings compared to the baseline method of distribution matching distillation method when applied to the CIFAR-10 dataset. Furthermore, we provide experimental evidence that Domain-Specific Variables (DSVs) offer unique insights beyond traditional distillation, and their integration leads to enhanced performance.

Conventional dataset distillation requires significant computational resources and assumes access to the entire dataset, an assumption impractical as it presumes all data resides on a central server. In this paper, we focus on dataset distillation in practical scenarios with access to only a fraction of the entire dataset. We introduce a novel distillation method that augments the conventional process by incorporating general model knowledge via the addition of Deep KKT (DKKT) loss. In practical settings, our approach showed improved performance compared to the baseline distribution matching distillation method on the CIFAR-10 dataset. Additionally, we present experimental evidence that Deep Support Vectors (DSVs) offer unique information to the original distillation, and their integration results in enhanced performance.
% \hh{Conventional dataset distillation needed huge computational cost and need the entire dataset, which is impractical as it assumes all data is in the central sever. In this paper, we focus on dataset distillation in practical scenarios where access to only a fraction of the entire dataset is available. We propose a new distillation method that enriches the conventional process by injecting general model knowledge through the addition of DKKT loss. In practical environments, our method demonstrated improved performance over the baseline distribution matching distillation method on the CIFAR-10 dataset. We also provide experimental evidence that DSVs contribute distinct information to the original distillation process, and their combination leads to enhanced performance.}

% The ABSTRACT is to be in fully justified italicized text, at the top of the left-hand column, below the author and affiliation information.
% Use the word ``Abstract'' as the title, in 12-point Times, boldface type, centered relative to the column, initially capitalized.
% The abstract is to be in 10-point, single-spaced type.
% Leave two blank lines after the Abstract, then begin the main text.
% Look at previous \confName abstracts to get a feel for style and length.
\end{abstract}    
\section{Introduction}
\label{sec:intro}
\begin{figure}[t]
  \centering
   \includegraphics[width=1.0\linewidth]{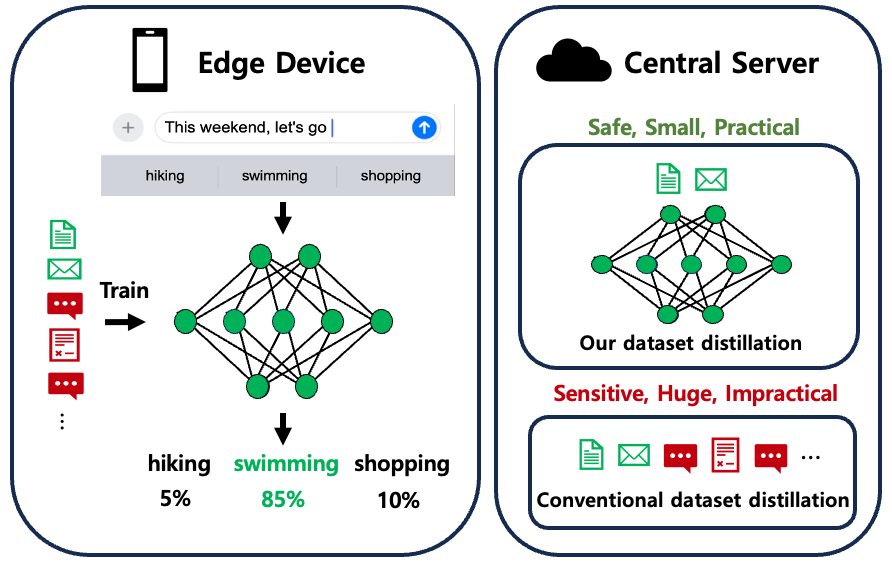}
   \caption{Dataset distillation in practical scenarios: data is gathered in edge devices, most of which is private (red), while safe data (green) that can be transferred to the central server is scarce. However a lightweight application model, continually trained on the entire dataset on the edge device, can be transferred to the server without any privacy concerns. }%Dataset distillation, conducted on the central server, can be performed safely and more quickly compared to conventional dataset distillation \nj{methods that require} the entire dataset. \nj{분량이 모자라면 이 그림은 빼도 될 것.}}
   \vspace{-6.0mm}
   \label{fig:data flow}
\end{figure}

With the emergence of large models such as Large Language Models (LLM)~\cite{achiam2023gpt4} and Diffusion models~\cite{betker2023dalle3,ideogram2023}, deep learning has been officially restructured into a data-centric paradigm.
%set-driven industry.
In this new era, data is considered a crucial asset and follows a specific data flow mechanism, as depicted in Figure.~\ref{fig:data flow}. 
%Data is typically generated and consumed \jh{in edge devices such as smartphones}
%as illustrated in Figure~\ref{fig:data flow}. 
Data is typically harvested from edge devices, where privacy and communication cost constraints prevent the transmission of entire datasets to a central server.
% Data is harvested from edge devices, where, due to concerns primarily related to privacy and communication issues, the practice is not to send the entire dataset to a central server. 
Instead, models are trained on edge devices such as personalized next word prediction or video recommendation are transferred to a central server. 

% This %Data-Driven Era 
% \jh{information age} has naturally fostered an interest in dataset distillation, aims to reduce computational and storge burdens. Normally, it relies two conditions. 
The information age has naturally fostered an interest in dataset distillation, aiming to reduce computational and storage burdens. Normally, it assumes two conditions. Firstly, most research employs bi-level optimization, thereby, huge computational cost. Secondly, it is assumed that the entire dataset is available for the distillation process. First condition suggests distillation is only feasible on central servers due to the computational burden.
% itself reveals a paradox. 
% \jh{This is because this condition worsens privacy concerns and increases communication burdens, where dataset distillation aims to mitigate.}
Additionally, it necessitates the transfer of extensive sensitive data from edge devices to central servers. This is paradoxical. While data distillation aims to mitigate the risk to privacy and reduce the burden of heavy communication, the training procedure itself introduces this very issue.

We propose a dataset distillation method for ``practical" real-world settings that existing dataset distillation approaches have not addressed. Our setting involves two conditions 1) During synthesizing, we cannot access whole dataset, we only used less than 1\% data. 2) We can utilize `single' model which is trained on edge devices. as edge devices normally trains model with gathered data for practical use (e.g. video suggestion, next-word-prompt). 
%More specifically, in these settings, we perform dataset distillation by utilizing a) single pretrained models and b) a small fraction of practical data (under 1$\%$).
%{To efficiently carry out distillation in these scenarios, it is crucial to extract general \nj{information about data} from the pretrained model while effectively condensing information from the ``practical dataset," which comprises the accessible portion of the entire training dataset.}
With this scenario, the performance of conventional algorithms declines sharply as the diversity of data diminishes due to smaller samples.
% In Figure~\ref{fig:pipc_accuracy}, the \nj{performance of }
% distribution matching (DM) \cite{DM}a
% \jh{dataset distillation declines sharply as the volume of available data shrinks, the diversity of data also diminishes, leading to a skewed representation in the distilled dataset.} 
\begin{figure*}[t]
  \centering
    \includegraphics[width=0.9\linewidth]{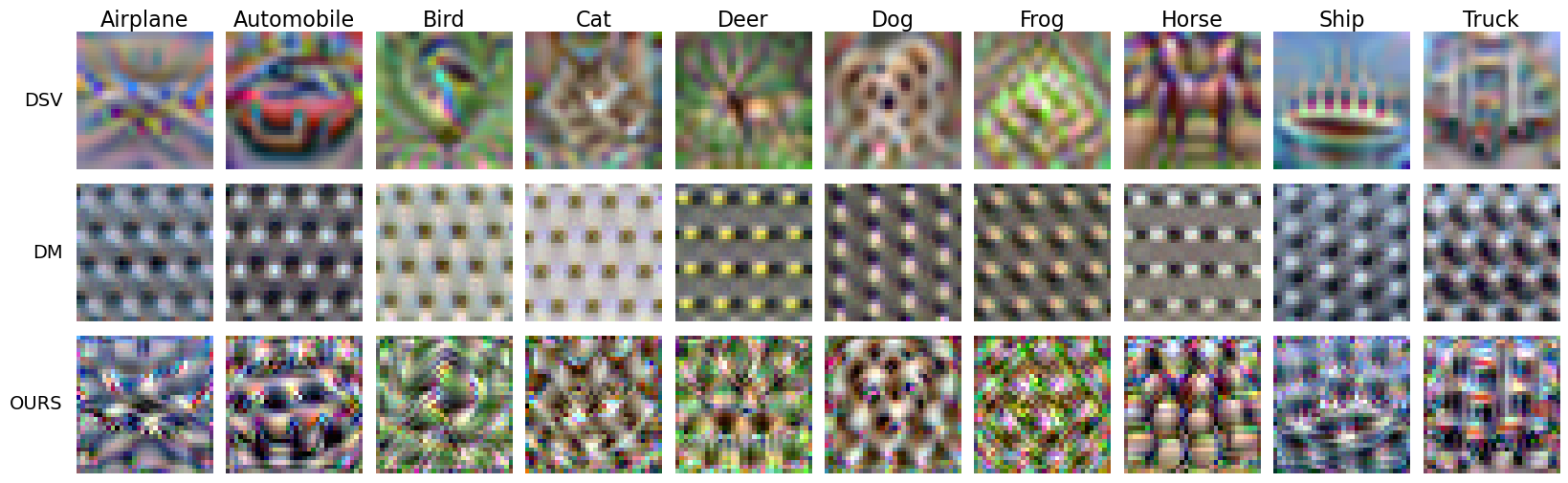}
   \caption{Qualitative results of our method, with 1 image per class (ipc) and 50 practically accessible images per class  (pipc).}
   \label{fig:qualitative results}
\end{figure*}
To address these challenges, we have integrated deep support vectors (DSVs) \cite{DSV}, a critical data feature extracted solely from pretrained models, into our data distillation process by DKKT loss. We achieved fast and effective distillation by merging DKKT loss with DM loss \cite{DM}, which narrows the distribution gap between training and synthetic data, eliminating the need for bilevel optimization. Our approach effectively distills data by extracting general and rich data features from pretrained models, improving performance even with limited practical data. 
% By extracting the DSV through inversion from pretrained models and utilizing it alongside the \nj{deep KKT (DKKT) loss \cite{DSV} --} which reduces the distribution gap between training and synthetic data -- we achieved fast and effective distillation without the need for bilevel optimization. 

% Notably in Figure~\ref{fig:pipc_accuracy}, the \ne{performance of distribution matching (DM) \cite{DM}} typically declines sharply as the available practical data decreases, in contrast to our method. 

Our contributions are summarized as follows:
\begin{itemize}
    \item To the best of our knowledge, we are the first to explore distillation using a "practical dataset" rather than the entire dataset, achieving performance improvements in low practical images per class settings compared to existing methods.
    \item We demonstrate that computationally unburdened distillation is feasible without bilevel optimization by simultaneously utilizing DKKT loss and DM loss.
    \item We propose a distillation approach that improves performance even in more practical scenarios where access to model weights is not available.
\end{itemize}

\section{Related Work }
 
\paragraph{Dataset Distillation.}
  Coreset selection, aimed at identifying representative images of the entire dataset, provides fair data representation but falls short in precisely mimicking the dataset distribution. The necessity arose not merely to select, but also to synthesize condensed images that more accurately reflect the overall training dataset, an approach referred to as dataset distillation. Most existing dataset distillation methods use bi-level optimization \cite{DD, DC}, or all the training data during the process of dataset distillation \cite{DM, CAFE}. In \cite{DDGAN}, it was argued that dataset distillation with synthesized data by GAN is beneficial from a privacy perspective, but they also used all the data to train the generative model for dataset distillation.

\paragraph{Model Inversion.}

Model Inversion tries to extract the data used during training from the trained model and Deep Support Vectors (DSV) \cite{DSV}, upon which our paper is based, can be considered one of the methods. In \cite{inversion_original}, inversion in visual models is defined through maximum bias. Subsequent researches \cite{inversion_maximummargin1, inversion_maximummargin2, Model_inversion_1} reconstruct the dataset by exploiting the property of a loss function with an exponential tail that maximizes the size of the boundary as training progresses and let the logit reflect this property.

% 본 논문에서는 practical한 상황에서의 dataset distillation에 대해서 다룬다. 기존의 dataset distillation의 경우 대부분 bi-level optimization을 사용하여서 train을 진행하거나 \cite{DD}, dataset distillation을 하는 과정 중에 모든 trained dataset을 사용하여서 학습을 진행하였다. \cite{DM, CAFE}. 이때 \cite{DDGAN}의 경우에는 dataset distillation을 할 때 있어서 GAN을 사용하여서 synthetized data로 dataset distillation을 하게 되면 privacy 관점에서 이점이 있다고 주장하였지만, 이 연구 또한 dataset distillation을 하기 위해서 그것을 위한 generation model을 데이터를 전부 사용하여 학습하였다는 특징이 있다.

% \paragraph{\jh{Model Inversion}}

% \jh{Model Inversion은 Trained model에서 실제로 사용된 dataset을 추출해내는 기법 중 하나로, 본 논문에서 중요하게 사용된 Deep Support Vectors (DSV) \cite{DSV}가 해당 분야 중 하나라고 볼 수 있다. 일반적으로 visual model에서의 inversion은 maximum bias를 통하여 정의된다 \cite{inversion_original} 이들은 학습이 진행되면 진행될수록 boundary의 크기를 maximize하는 expotential tail을 가진 loss 함수의 특성을 활용하여, logit이 이 특성을 반영하도록 하여서 dataset을 reconstruction한다. \cite{inversion_maximummargin1, inversion_maximummargin2, Model_inversion_1}}

%  \begin{figure}[t]
%   \centering
%    \includegraphics[width=0.7\linewidth]{./figures/pipc_acc.png}
%    \caption{Accuracy for different pipc setups for ipc 1}
%    \label{fig:pipc_accuracy}
% \end{figure}

% \
\section{Method}
% jku
\label{sec:method}
\paragraph{Notation.} In this section, we explore methodologies for efficient data condensation in practical scenarios characterized by constrained accessibility to the training dataset without a significant computational burden. Consider a scenario wherein access is restricted to a subset $\mathcal{T}'=\{(x_i,y_i)\}$ of the complete training set $\mathcal{T}$, such that $\mathcal{T}'\subset\mathcal{T}$, and a model $\Phi(\cdot; \theta)$ has been trained on $\mathcal{T}$. The cardinality $|\{(x_i,y_i)|y_i=c\}|$ for class $c$ within $\mathcal{T}'$ is defined as the practically accessible images per class, denoted as \textbf{\textit{pipc}}. Similarly for class $c$, the cardinality of the synthetic dataset $\mathcal{S}=\{(s_i,y_i)\}$, expressed as $|\{(s_i,y_i)|y_i=c\}|$, is defined as the learnable images per class, or \textbf{\textit{ipc}}. Our goal is to create a synthesized dataset $\mathcal{S}$ that can achieve the performance of the entire training dataset $\mathcal{T}$ only utilizing $\mathcal{T}'$. 

% \hh{ 이번 섹션에서는 training dataset에 대한 접근이 제한적인 환경에서 high computation load 없이 효과적으로 condensation을 할 수 있는 방법에 대하여 논한다. 우리는 class $\mathcal{C}$인 전체 training set $\mathcal{T}$에 대하여 일부의 데이터셋 $\mathcal{T}'=\{(x_i,y_i)\}$,$\mathcal{T}'\subset\mathcal{T}$에만 접근이 가능하며 , $\mathcal{T}$에 대하여 pretrained 된 모델 $\Phi(\cdot; \theta)$를 가지고 있다고 하자. class $c=y_i$에 해당하는 $\mathcal{T}'$의 cardinality $|\{(x_i,y_i)|y_i=c\}|$ 를 class 별 practically accessable image per class: \textit{pipc}라고 정의하며, class c에 해당하는 synthetic dataset $\mathcal{S}=\{(s_i,y_i)\}$의 cardinality $|\{(s_i,y_i)|y_i=c\}|$ 를 class 별 학습가능 image(s) per class \textit{ipc}로 정의한다. 이때 우리의 목표는 $\mathcal{T}'$만을 가지고 전체 training dataset $\mathcal{T}$의 성능을 낼 수 있는 synthesized dataset $\mathcal{S}$ 를 만드는 것이다.} \jh{이거 우리가 처음 한다 이게 어떤 의미인지?}

\subsection{Preliminaries}

\begin{table*}[t]
\centering
\begin{tabular}{c|cc|ccccc}
\toprule
Img/Cls             & Practical Img/Cls & ratio \% & DM                          & DM$^*$     & DSV (noise)                       & DSV (real)                          & Ours               \\ \hline
\multirow{3}{*}{1}  & 10                & 0.2      &{22.42$\pm$0.43}             & {23.26$\pm$0.66}    &\multirow{3}{*}{25.3$\pm$0.5}     & \multirow{3}{*}{16.68$\pm$0.49}    &{25.53$\pm$0.45}    \\
                    & 50                & 1        &{24.42$\pm$0.29}             &{25.83$\pm$1.62}     &                                  &                                    &{28.20$\pm$0.69}     \\
                    & all               & 100      &{26.39$\pm$0.98}             &{27.08$\pm$0.99}     &                                  &                                    &{29.95$\pm$0.23}     \\\hline
\multirow{3}{*}{3}  & 10                & 0.2      &{28.44$\pm$0.30}             &{26.04$\pm$0.29}     & \multirow{3}{*}{23.98$\pm$0.21}  & \multirow{3}{*}{27.95$\pm$0.61}    &{30.38$\pm$0.47}      \\
                    & 50                & 1        &{34.49$\pm$0.65}             &{34.26$\pm$0.65}     &                                  &                                    &{37.41$\pm$0.72}     \\
                    & all               & 100      &{39.18$\pm$0.46}             &{37.37$\pm$0.42}     &                                  &                                    &{39.42$\pm$0.37}     \\ \hline
\multirow{3}{*}{10} & 10                & 0.2      &{31.13$\pm$0.45}             &{29.77$\pm$0.66}     &\multirow{3}{*}{24.88$\pm$1.29}   &\multirow{3}{*}{36.39$\pm$0.52}     &{31.55$\pm$0.72}     \\
                    & 50                & 1        &{44.48$\pm$0.43}             &{41.27$\pm$0.59}     &                                  &                                    &{44.51$\pm$0.38}     \\
                    & all               & 100      &{49.75$\pm$0.36}             &{45.98$\pm$0.53}     &                                  &                                    &{47.38$\pm$0.93}     \\ \hline
\multirow{2}{*}{50} & 50                & 1        &{49.44$\pm$0.56}             &{48.05$\pm$0.42}     &\multirow{2}{*}{23.68$\pm$1.35}   &\multirow{2}{*}{47.01$\pm$0.80}     &{51.12$\pm$0.79}     \\
                    & all               & 100      &{60.59$\pm$0.41}             &{56.34$\pm$0.56}     &                                  &                                    &{57.62$\pm$0.33}     \\ \bottomrule

\end{tabular}
\caption{Accuracy for CIFAR 10 dataset in different pipc setup. For DSV, ``noise" indicates that initialization starts from noise, while ``real" indicates starting from a real image. DM$^*$ indicate DM implimented in our code. }
\label{tab:pipc}
\end{table*}

\paragraph{Deep Support Vectors.}

Support Vector Machines (SVMs) are recognized as one of the effective tools for analyzing linearly separable data, where the support vectors are commonly known to satisfy the Karush-Kuhn-Tucker (KKT) conditions. In \cite{DSV}, the concept of KKT conditions that the support vectors must satisfy was extended to deep learning models, creating Deep Support Vectors (DSVs) that fulfill a similar role within deep learning frameworks as support vectors do in SVMs. The KKT conditions for DSV candidates $\{x_i\}$ and corresponding Lagrange multipliers $\lambda_i$ were redefined in terms of a model $\Phi(x;\theta)$ and a loss function $\mathcal{L}(\Phi(x_i; \theta), y_i)$ as presented below: \\
{\small
\begin{equation} \label{eq:DeepKKT}
% \begin{small}
\begin{aligned}
&\text{Primal feasibility:} && \forall\, i \in \mathcal{I}, \quad \argmax_c\; \Phi_c (x_i; \theta^*) = y_i \\ 
&\text{Dual feasibility:} && \forall\, i \in \mathcal{I}, \quad \lambda_i \geq 0, \\
&\text{Stationarity:} && \theta^* = -\sum_{i=1}^n \lambda_i \nabla_{\theta} \mathcal{L}(\Phi(x_i; \theta^*), y_i), \\
&\text{Manifold:} && \forall\, i \in \mathcal{I}, \quad x_i \in \mathcal{M}. 
\end{aligned}
% \end{small}
\end{equation}
}

DSVs, akin to conventional support vectors in containing information about the decision boundary, also embody representations of the dataset itself, thus possess the potential to play a significant role in dataset condensation. A significant advantage of this approach is the ability to generate DSVs using only a pretrained model, without direct access to the training dataset.
% \hh{SVM은 linearly seperable data들을 해석하는 좋은 도구중 하나로 알려져있으며, 일반적으로 support vector들은 KKT condition을 만족하는것으로 알려져있다. Lee et al \cite{DSV}은 support vector들이 만족해야할 KKTcondition을 딥러닝 모델에 적용하여, SVM에서 support vector들이 하는 역할을 딥러닝 모델에서 동일하게하는 Dsv들을 만들었다. Dataset $\{x_i,y_i\}$와 이에 대응되는 Lagrange multipliers $\lambda_i$ 들에 대한 KKT condition을 model $\Phi(\theta)$, loss function $\mathcal{L}(\Phi(x_i; \theta), y_i)$에 대하여 Eq.~(\ref{eq:DeepKKT})에서 새로히 정의 하였다. DSV들은  conventional SV와 동일하게 decision boundary에 대한 정보를 가지고 있기도 하지만 dataset 자체의 representation이기도 하므로, dataset condensation에서의 role 또한 수행할 수 있다. Training dataset에 대한 접근이 아예 없이 기학습된 모델만 가지고도 생성할 수 있다는 장점이 있다.} 

% \nj{notation에 대한 설명 필요.}

%  \begin{figure}[t]
%   \centering
%    \includegraphics[width=0.8\linewidth]{./figures/data_manifold.png}
%    \caption{Explanation for overall method    설명 추가noise(random) 추가 ipc1일때 이런 설명 추가}
%    \label{fig:data_manifold}
% \end{figure}

\paragraph{Distribution Matching.}
Distribution matching (DM) \cite{DM} proposes a novel dataset distillation method that neither employs gradient matching nor updates model parameters. This method synthesizes images efficiently by comparing embeddings in lower dimensions, utilizing an arbitrary embedding function $\psi$. It accomplishes this by minimizing the classwise Maximum Mean Discrepancy (MMD) between the real dataset $\mathcal{T}$ and the synthetic dataset $\mathcal{S}$, thereby reducing the distance between the two distributions in the embedding space. This efficient algorithm avoids bi-level optimization but, unlike other optimization methodologies \cite{DC, DD}, it cannot utilize information from trained models.
\subsection{Combining knowledge to mitigate low diversity}
\paragraph{Problem.}
In practical scenarios, access to the entire training dataset is often restricted due to concerns related to communication burden and data privacy, resulting in availability to merely a fraction (approximately 1$\%$) of the total dataset. This significant reduction in dataset size inevitably diminishes the diversity within the dataset, which in turn, can compromise the effectiveness of dataset distillation processes.

% To mitigate these limitations and enhance the quality of distillation, the integration of DKKT Loss with the DM loss is proposed. This strategy leverages the generalized information previously extracted by the pretrained model, thereby augmenting the distillation process with additional, valuable insights.}
% \hh{In realistic scenarios, we have limited access to the total training dataset because of communication and privacy issues; therefore 가지고 있다 only a small portion(1$\%$). 크기가 줄어든 dataset diversity is reduced leading to inaccurate distillation.
% Therefore we add a DKKT LOSS to the DM loss to compensate additional general information extracted in the pretrained model to reinforce distillation quality}

% \paragraph{Distribution Matching Loss (Data Knowledge Loss)}
\paragraph{DM loss (Data knowledge loss).}
% \jh{이 텀은 이래서 Data knowledge}
% \hh{  $L_{\text{DKKT}}$는 기학습된 모델만을 이용하지만 이는 DSV들을 실제 data distribution으로 보내는데에는 한계가 있다. 따라서 computation을 최소화하면서 DSV들을 좋은 initial point로 하여 synthesized 된 이미지들을 data manifold에 가져다 놓을 필요성이 있다. 우리는 Distribution matching loss를 활용하여, 접근이 가능한 training datset $\mathcal{T}'={(x_i,y_i}$을 anchor로 삼아 효과적으로 synthesized image에 실제 data 정보들을 주입하였다. $L_{\text{DM}}$은 아래 Eq.~\ref{eq:DMloss}로 적용했으며 randomly initialized embedding function $\psi_\theta$은 ConvNet을 사용하였고, augmentation factor $\omega$는 random하게 뽑아서 적용하였다.} 

We utilize the DM loss to tackle the significant computational costs associated with bi-level optimization, which becomes increasingly problematic as data resolution expands. The DM loss effectively integrates the knowledge and characteristics of the accessible training dataset $\mathcal{T}'=\{(x_i,y_i)\}$ into the synthesized data $\mathcal{S}$, bypassing the need for bi-level optimization, as demonstrated below: 
{\small
\begin{equation}\label{eq:DMloss}
%\begin{footnotesize}
% \small
L_{\text{DM}}(\mathcal{S}) = \left\| \frac{1}{|\mathcal{T}'|} \sum_{i=1}^{|\mathcal{T}'|} \psi(\mathbf{A}(x_i)) - \frac{1}{|\mathcal{S}|} \sum_{j=1}^{|\mathcal{S}|} \psi(\mathbf{A}(s_j)) \right\|^2
%\end{footnotesize}
\end{equation}
}
The implementation of $L_{\text{DM}}$ employs a ConvNet\cite{convnet} for the randomly initialized embedding function $\psi$. Furthermore, to effectively embed prior knowledge into the synthesized images, a differentiable Siamese augmentation \cite{DSA} $\mathbf{A}(x_i)$ is applied.
% Utilizing only a pretrained model, the DKKT loss encounters limitations in accurately mapping DSVs to the actual data distribution. Consequently, there is a need to position the synthesized images on the data manifold using DSVs as good initial points, while minimizing computation. We employed the distribution matching loss, using the accessible training dataset $\mathcal{T}'=\{(x_i,y_i)\}$ as an anchor, to effectively infuse real data characteristics into the synthesized images. 
% $L_{\text{DM}}$ (Eq.~\ref{eq:DMloss}) was carried out by utilizing ConvNet for the randomly initialized embedding function $\psi_\theta$, and the augmentation factor $\omega$ was selected randomly.

\paragraph{DKKT loss (Model knowledge loss).}
To address the issue of reduced diversity due to limited data availability, we adopt the DKKT loss. This term derives knowledge exclusively from pretrained models, without accessing actual data.
The DKKT loss is composed of two terms: the primal loss, which guarantees the accurate classification of DSVs into their respective classes, and the stationarity loss, aimed at facilitating the convergence of these vectors towards the support vectors situated near the decision boundary. For a set of DSV candidate and the corresponding Lagrange multiplier pairs $\mathcal{X}=\{({x_i},\lambda_i)\}_{i=1}^n$, %alongside their associated Lagrange multipliers $\lambda_i$, and considering a specified multiclassification loss $L$, 
the primal and stationarity losses that reflect Eq.~\ref{eq:DeepKKT} are defined as follows:
% \hh{ DKKT loss는 DSV들을 올바른 class로 classify하는 primal loss와, decision boundary에 있는 support vector로 수렴하게 하는 stationarity condition으로 구성된다. DSV candidate $x_i$들과 이에 corresponding하는 Lagrange multipliers $\lambda_i$ 들에 그리고 for some multiclassification loss $L_i$ 로부터 primal loss 와 stationarity loss는 다음과 같이 표현될 수 있다. 
%
\begin{equation}\label{eq:total_loss}
\begin{split}
&L_{\text{primal}}(\mathcal{X}) = \frac{1}{n} \sum_{i=1}^n L(\Phi(x_i; \theta^*), y_i), \\
%\end{equation}
%
%
%\begin{equation}\label{eq:stationarity}
&L_{\text{stat}}(\mathcal{X}) = D(\theta^*, -\sum_{i=1}^n \lambda_i \nabla_\theta L(\Phi(x_i;\theta^*), y_i)).
\end{split}
\end{equation}

Here, $D$ is some distance metric and $L$ is a multiclassification loss. The DKKT loss for a synthetic dataset $\mathcal{S}$ can be represented as a linear combination of the primal loss and the stationarity loss, modulated by a weighting factor $\alpha$:
\begin{equation}\label{eq:DKKT loss}
L_{\text{DKKT}}(\mathcal{S}) = L_{\text{primal}}(\mathcal{S}) + \alpha L_{\text{stat}}(\mathcal{S}).
\end{equation}

\paragraph{Practical Dataset Distillation.}
Both $L_{\text{DKKT}}(\mathcal{S})$ and $L_{\text{DM}}(\mathcal{S})$ rely solely on the synthetic dataset $\mathcal{S}$, allowing for the simultaneous application of gradient descent through their linear combination. The final total loss $L_{\text{Total}}(\mathcal{S})$ is represented as a linear combination of these two losses, using weighting factor $\gamma$, as delineated below:
% \hh{ $L_{\text{DKKT}}(S)$ 와 $L_{\text{DM}}(S)$ 모두 synthetic dataset $S$에만 의존하는 식으로, 두 loss를 선형 결합하여 동시에 gradient decent를 할 수 있다. 두 loss를 weighting factor $\gamma_1$와 $\gamma_2$로 선형결합하여 최종 total loss $L_{\text{Total}}(S)$를 아래 Eq.~\ref{eq:Totalloss}로 표현하였다.}
%
\begin{equation}\label{eq:Totalloss}
L_{\text{Total}}(\mathcal{S}) = L_{\text{DKKT}}(\mathcal{S}) +\gamma L_{\text{DM}}(\mathcal{S}).
\end{equation}

\section{Experiments}
\label{sec:experiments}
% \paragraph{Experimental Settings.}

% \hh{  Our classification performance was evaluated on the CIFAR-10 \cite{dataset:cifar10} dataset using a three-layer ConvNet\cite{convnet} structure within the data condensation task. We conducted experiments across ipc values of 1/3/10/50 and pipc settings of 10/50/all. During dataset synthesis, for pipc 10/50/all, we utilized stationarity rate $\alpha$ at 0.1, 0.01, and 0.001 respectively and set DM ratio $\gamma$ at 0.01 for ipc 50 with 0.001 applied in other instances. Additionally for ipc 1, initialization was done with noise, whereas for ipc 3/10/50, initialization was conducted with real images. During evaluation, the SAM optimizer \cite{SAM} was utilized with a learning rate of 0.1 and $\rho$ of 0.001 across 5000 epochs.}
% \hh{우리는 기본적으로 CIFAR 10 dataset \cite{dataset:cifar10}에서 실험을 진행하였으며 data condensation task에 많이 사용되는 Convnet\cite{convnet} stucture을 모델로 evaluation 하였다. IPC 1/3/10/50에 대하여 pipc 10/50/all으로 실험을 하였으며, dataset synthesis일때는 pipc 10/50/all 에 대해서 각각 $\gamma_1$(DKKT ratio) 0.1/0.01/0.001을 사용하였으며, $\gamma_2$ (DM ratio)는 ipc 50에서는 0.01, 그 외의 경우에는 0.001을 사용하였다. 또한 ipc1에 대해서는 init with noise로 했으며, ipc 3/10/50에 대해서는 init with real images 로 하였다.  evaluation에서는 SAM optimizer로 lr:0.1, rho:0.001로 5000 epoch로 실험하였으며 it took 00 minutes}

% \paragraph{DSVs Encompass Rich and Important Information.}
\paragraph{DM and DSVs on practical settings.}
% \hh{Table~\ref{fig:data flow} 에서 practical settings where pipc ratio is same or less then 1$\%$ 에서 ours가 DM과 DSV에 비해 성능향상이 있음을 확인하였다. DSV의 경우ipc 1인 세팅에서는 DM에 비해서 성능이 좋았지만, ipc3 이후부터는 DM에 비해 성능이 떨어짐을 확인할 수 있다. 이는 DsV가 training dataset자체에 대한 접근이 없어 prior에 대한 정보를 효과적으로 주입하지 못하였으며, support vector 특성상 decision boundary에 있는 data에 편향되기 때문에 mode collapse가 발생했기 때문이라고 볼 수 있다. DM method는 ipc와 pipc가 높아 질수록 성능이 확실히 잘 올라갔는데, 이는 가용할 수 있는 training set이 많아질수록 synthetic set에 저장할 수 있는 정보량이 많아지기 때문이다. 하지만 ipc가 1과 같이 작은 경우 synthetic data 자체가 전체 $\mathcal{T}$의 unbiased estimater이 되므로 Fig~\ref{fig:qualitative results}와 같이 blurry averaged feature들이 나오며, low pipc에서도 소수의 training data들에 bias가 되기 때문에 성능이 감소하였다고 할 수 있다.
%  ipc 1 pipc 50이하인 setting에서, DM은 real data에서 부터 시작했음에도 불구하고 training dataset이 없는 noise에서 시작한 DSV보다도 성능이 낮게 나오는데 이는 DSV에게 소수의 real data에게 없는 rich information을 함유하고 있음을 알 수 있다. DSVs들은 기학습된 classifier model에서 뽑아 냄으로서 data들을 판단할때에 있어서 중요하고 general한 feature들을 담았기 때문에, DM을 적용하는 data들과 다르고 중요한 information을 담았다. 따라서 dsv는 데이터셋에 접근이 적은 practical settings에서 DM이 줄 수 없는 중요한 정보들을 함유하고 있다는 점에서 DKKT loss는 매우 crucial 한 역할을 한다고 할 수 있다.}
In Table~\ref{fig:data flow}, our method showed enhanced performance compared to DM and DSV in scenarios with a pipc ratio of 1$\%$ or lower. Notably, DSV surpassed DM for ipc 1 setting but its advantage waned at ipc values beyond 3, a trend potentially linked to DSV's restricted training dataset access. Additionally, 
%This limitation likely hampered the integration of prior knowledge and introduced 
DSV has a bias towards decision boundary, resulting in limited diversity. 
Conversely, DM experienced progressive performance boosts as ipc and pipc increase, benefiting from the increased capacity of the expanded synthetic set to hold more information from an increased volume of training data. At minimal ipc values such as 1, synthetic data became an unbiased estimator for the entire dataset $\mathcal{T}$, yielding blurred, generalized features as depicted in Fig.~\ref{fig:qualitative results} and diminishing performance due to a lean towards a scant number of training examples in scenarios with low pipc.

\paragraph{DSVs encompass global information.}
In settings with ipc 1 and pipc ≤ 50, DM performed worse than DSVs started from random noise. This reveals DSVs hold rich information not found in limited real data. DSVs, extracted from a pretrained classifier, contain crucial and broad features important for data evaluation, embedding significant insights not seen in DM-utilized data. This is because DM does not use bi-level optimization and therefore cannot access model knowledge.
Therefore, in situations with limited dataset access, DSVs offer critical information that DM cannot provide, highlighting the essential role of DKKT loss in these scenarios.
% + dm에서 bilevel optimization역할을 dsv가 해준다
  % ipc 높아질수록 dm loss많이준 이유 추가

\paragraph{DSV and DM encapsulate different information.} 
\label{sec:discussion}
\begin{figure}[t]
  \centering    \includegraphics[width=0.9\linewidth]{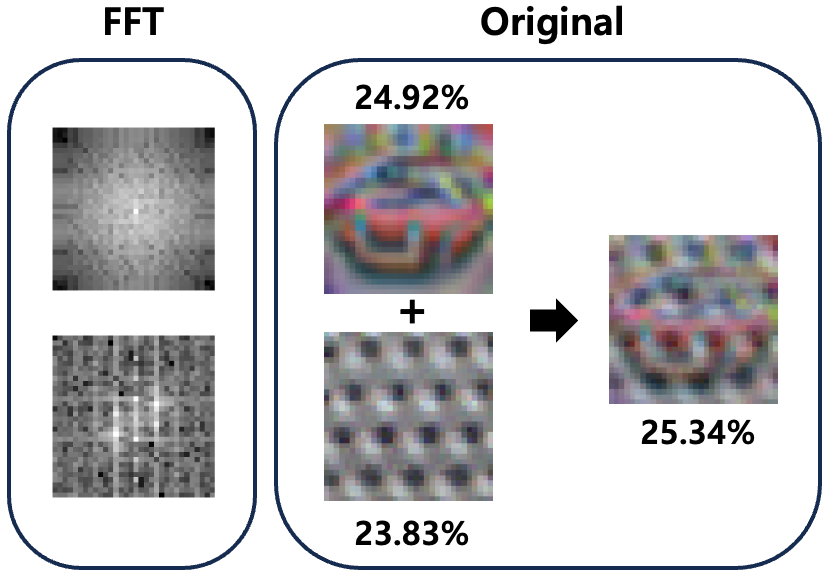}
  \vspace{-0.5mm}
   \caption{Results for the average of DSVs (top) and DM (bottom). The Fourier-Transformed images (FFT) indicate that the two lie in different frequency domains and the averaged sample resulted in better accuracy (25.34\%). }
  \vspace{5.0mm}
   \label{fig:superposition}
\end{figure}

Fig.~\ref{fig:superposition} presents a strategy for enhancing performance in practical scenarios where access to the model is restricted, but pre-extracted DSVs are available. Evaluating averaged images, created by blending pre-extracted DSVs with images synthesized via DM, resulted in performance improvements. In the figure, the synthesized image with DM is obtained using 50 pipc.
Conventionally, data condensation does not allow for the mixing of information from two synthetic data sets. However, as observed in Fig.~\ref{fig:qualitative results}, despite being updated simultaneously by the loss, the information appears to be combined. This can be attributed to DSVs predominantly containing low-frequency information (Fig.~\ref{fig:superposition}.left.top), while DM-derived images tend to have a higher frequency content (Fig.~\ref{fig:superposition}.left.bottom), allowing for minimal information loss when simply combining the two images.
% \nj{appendix에 식 2를 이용해서 DSV/DM/Ours의 distribution matching score를 비교하면 좋을 것 같음. 공간이 있으면 여기에 글로 표현하든지. }
% \hh{ Fig~\ref{fig:superposition}에서는 practical setting에서 model에 대한 접근이 불가하고 model에서 사전에 뽑아놓은 dSV들에 대한 접근만 가능한 시나리오에서 성능을 높이는 방안을 제시한다. 사전에 뽑아 놓은 DSV들과 DM으로 뽑은 이미지들에 대하여 $\lambda$ 비율을 활용하여 superposition한 이미지를 evaluation했더니 성능 향상이 있었다. 일반적으로 data condensation이라고 하면 synthetic data 두 세트에서 두개의 정보를 혼합할 수 있는 방법은 없다. 하지만 Fig~\ref{fig:qualitative results}를 보면 loss로 한꺼번에 updated 됐음에도 두 정보가 superpositioned 된것처럼 보인다. 이는 DSV에서는 low frequency정보가 주를 이루는 반면에, DM에서는 high frequncy이 비교적 많아, 두 이미지를 단순히 더해도 정보손실이 비교적 적게 나기 때문이라고 할 수 있다.}

\paragraph{Qualitative analysis.}
% \hh{  Fig~\ref{fig:qualitative results}에서는 ipc1 pipc 50에서의 DSV, DM, OURS의 qualitative results를 나타낸다. DSV들은 CIFAR10 class 들에 대해서 shape과 color feature들이  사람이 구분할 정도로 확연하게 나타나지만, DM에서는 class별 대표 색만 알아볼 수 있을만한 격자무늬 이미지들이 생성되었음을 알 수 있다. DM 이미지에서는 class당 50장의 정보들을 1장에 cram해 넣어야하기 때문에 class별 대표 중요 feature들을 넣기 보다는 averaging된 정보들이 들어가있다. 두 방법의 loss를 섞은 방법에서는 이미지상으로 dSV와 DM의 이미지들이 superposition 된것으로 보이는데, 이는 dSV의 model정보는 유지한채 practical dataset과의 distribution matching에도 성공했음을 추측할 수 있다.}
In Fig.~\ref{fig:qualitative results}, qualitative results are presented for ipc 1 and pipc 50 settings within the CIFAR-10 dataset. DSVs clearly display shape and color features that are discernible by humans. In contrast, DM-produced images primarily contain patterns indicative of class-specific colors. Due to DM's task of condensing information from 50 images into a single image, the resulting visuals tend to feature generalized data rather than unique, class-specific attributes. Our method which combines the two losses, results in images that appear to be a fusion of DSV and DM elements. This indicates that while maintaining the model information from DSVs, the approach also effectively matches the distribution of the practical dataset. Due to diverse sources of information, the integration of the two methods could be interpreted as forming an ensemble.

\section{Conclusion}
\label{sec:conclusion}
In this paper, we present a new dataset distillation method designed for scenarios where only a small fraction of the dataset is accessible, due to communication constraints and privacy issues. To overcome the information shortfall from such limited data, we enhance the model's knowledge by merging DM loss (data knowledge) with DKKT loss (model knowledge). Our results indicate improved accuracy in practical environments compared to conventional DM or DSV approaches. This enhancement arises from integration of diverse sources of information, operating as an ensemble.
%specific knowledge provided by each method, collectively operating as an ensemble.}
\newpage
% \hh{In this paper, we propose a new dataset distillation method in practical settings: settings where only a small portion of the dataset is available due to communication burden and privacy. We solve the lack of information due to small dataset by injecting model knowledge, i.e. combining DM loss(data knowledge) with an additional DKKT loss(general model knowledge). Experiments show that this method shows better accuracy in low pipc settings than base DM or DSVs. This is because two methods hold different information therefore results as an ensemble when combining these.}
% \hh{ practical한 환경에서 잘된다 등등}
% \hh{우리의 contribution recap
% 모델의 정보를 효과적으로 dm방법과 함께 녹여냄 -> model 정보를 + practical dataset 정보 가 가능함
% }
% \input{sec/02_formatting}
% \input{sec/03_finalcopy}
{
    \small
    \bibliographystyle{ieeenat_fullname}
    \bibliography{main}
}

\clearpage
\setcounter{page}{1}
\maketitlesupplementary
\appendix
\section{Experimental Settings.}
\label{sec:experimental_settings}
Our classification performance was evaluated on the CIFAR-10 \cite{dataset:cifar10} dataset using a three-layer ConvNet\cite{convnet} structure within the data condensation task. We conducted experiments across ipc values of 1/3/10/50 and pipc settings of 10/50/all. During dataset synthesis, for pipc 10/50/all, we utilized stationarity rate $\alpha$ at 0.1, 0.01, and 0.001 respectively and set DM ratio $\gamma$ at 0.01 for ipc 50 with 0.001 applied in other instances. Additionally for ipc 1, initialization was done with noise, whereas for ipc 3/10/50, initialization was conducted with real images. During evaluation, the SAM optimizer \cite{SAM} was utilized with a learning rate of 0.1 and $\rho$ of 0.001 across 5000 epochs.
% \section{Rationale}
% \label{sec:rationale}
% % 
% Having the supplementary compiled together with the main paper means that:
% % 
% \begin{itemize}
% \item The supplementary can back-reference sections of the main paper, for example, we can refer to \cref{sec:intro};
% \item The main paper can forward reference sub-sections within the supplementary explicitly (e.g. referring to a particular experiment); 
% \item When submitted to arXiv, the supplementary will already included at the end of the paper.
% \end{itemize}
% % 
% To split the supplementary pages from the main paper, you can use \href{https://support.apple.com/en-ca/guide/preview/prvw11793/mac#:~:text=Delete%20a%20page%20from%20a,or%20choose%20Edit%20%3E%20Delete).}{Preview (on macOS)}, \href{https://www.adobe.com/acrobat/how-to/delete-pages-from-pdf.html#:~:text=Choose%20%E2%80%9CTools%E2%80%9D%20%3E%20%E2%80%9COrganize,or%20pages%20from%20the%20file.}{Adobe Acrobat} (on all OSs), as well as \href{https://superuser.com/questions/517986/is-it-possible-to-delete-some-pages-of-a-pdf-document}{command line tools}.

\end{document}